\pdfoutput=1  

\documentclass[11pt]{article}

\usepackage[margin=1in]{geometry}
\usepackage{amsmath,amssymb,amsfonts}
\usepackage{algorithm}
\usepackage{algorithmic}
\usepackage{graphicx}
\usepackage{booktabs}
\usepackage{microtype}
\usepackage{xcolor}
\usepackage{authblk}
\usepackage{hyperref}
\hypersetup{
  colorlinks=true,
  linkcolor=blue,
  citecolor=blue,
  urlcolor=blue
}

\usepackage{bibentry}

\usepackage{subcaption}
\usepackage{amsthm}

\usepackage{amsmath}
\usepackage{amssymb}
\usepackage{mathtools}
\usepackage{bm}

\usepackage{makecell}    
\usepackage{arydshln}    
\usepackage{array}       
\usepackage{graphicx}
\usepackage{CJKutf8}
\usepackage{float}

\usepackage{listings}
\usepackage{xcolor}
\usepackage[T1]{fontenc}
\usepackage[utf8]{inputenc}

\definecolor{mygray}{rgb}{0.5,0.5,0.5}
\definecolor{myorange}{rgb}{0.8,0.3,0}
\definecolor{myblue}{rgb}{0.2,0.3,0.8}

\lstdefinelanguage{jsonld}{
  morestring=[b]",
  morecomment=[l]{//},
  morekeywords={true,false,null},
  sensitive=false,
  literate=
   *{:@}{{{\textcolor{mygray}{:}}@}}1
    {{"@id"}}{{{\textcolor{myblue}{"@id"}}}}1
    {{"@type"}}{{{\textcolor{myblue}{"@type"}}}}1
    {{"rdfs:label"}}{{{\textcolor{myblue}{"rdfs:label"}}}}1
    {{"rdfs:comment"}}{{{\textcolor{myblue}{"rdfs:comment"}}}}1
    {{"rdfs:subClassOf"}}{{{\textcolor{myblue}{"rdfs:subClassOf"}}}}1
    {{"rdfs:domain"}}{{{\textcolor{myblue}{"rdfs:domain"}}}}1
    {{"rdfs:range"}}{{{\textcolor{myblue}{"rdfs:range"}}}}1
    {{"rdf:Property"}}{{{\textcolor{myblue}{"rdf:Property"}}}}1
    {{"owl:Class"}}{{{\textcolor{myblue}{"owl:Class"}}}}1
    {{"appliesNorm"}}{{{\textcolor{myblue}{"appliesNorm"}}}}1
    {{"toFact"}}{{{\textcolor{myblue}{"toFact"}}}}1
    {:}{{\textcolor{mygray}{:}}}1
    {,}{{\textcolor{mygray}{,}}}1
}

\usepackage{tcolorbox}
\tcbuselibrary{skins, breakable}
\usepackage{caption}

\usepackage[numbers]{natbib}
\title{Capturing Legal Reasoning Paths from Facts to Law in Court Judgments using Knowledge Graphs}

\author[1,2]{Ryoma Kondo\thanks{\texttt{kondor@g.ecc.u-tokyo.ac.jp}}}

\author[3]{Riona Matsuoka}

\author[3]{Takahiro Yoshida}

\author[4]{Kazuyuki Yamasawa}

\author[1,2]{Ryohei Hisano\thanks{\texttt{hisanor@g.ecc.u-tokyo.ac.jp}}}

\affil[1]{Graduate School of Information Science and Technology, The University of Tokyo}

\affil[2]{The Canon Institute for Global Studies}

\affil[3]{Graduate Schools for Law and Politics, The University of Tokyo}

\affil[4]{TKC Corporation}

\begin{document}

\maketitle

\begin{abstract}
Court judgments reveal how legal rules have been interpreted and applied to facts, providing a foundation for understanding structured legal reasoning. However, existing automated approaches for capturing legal reasoning, including large language models, often fail to identify the relevant legal context, do not accurately trace how facts relate to legal norms, and may misrepresent the layered structure of judicial reasoning. These limitations hinder the ability to capture how courts apply the law to facts in practice. In this paper, we address these challenges by constructing a legal knowledge graph from 648 Japanese administrative court decisions. Our method extracts components of legal reasoning using prompt-based large language models, normalizes references to legal provisions, and links facts, norms, and legal applications through an ontology of legal inference. The resulting graph captures the full structure of legal reasoning as it appears in real court decisions, making implicit reasoning explicit and machine-readable. We evaluate our system using expert annotated data, and find that it achieves more accurate retrieval of relevant legal provisions from facts than large language model baselines and retrieval-augmented methods.

\end{abstract}

\section{Introduction}

\begin{figure*}[t]
  \centering
  \includegraphics[width=1\textwidth]{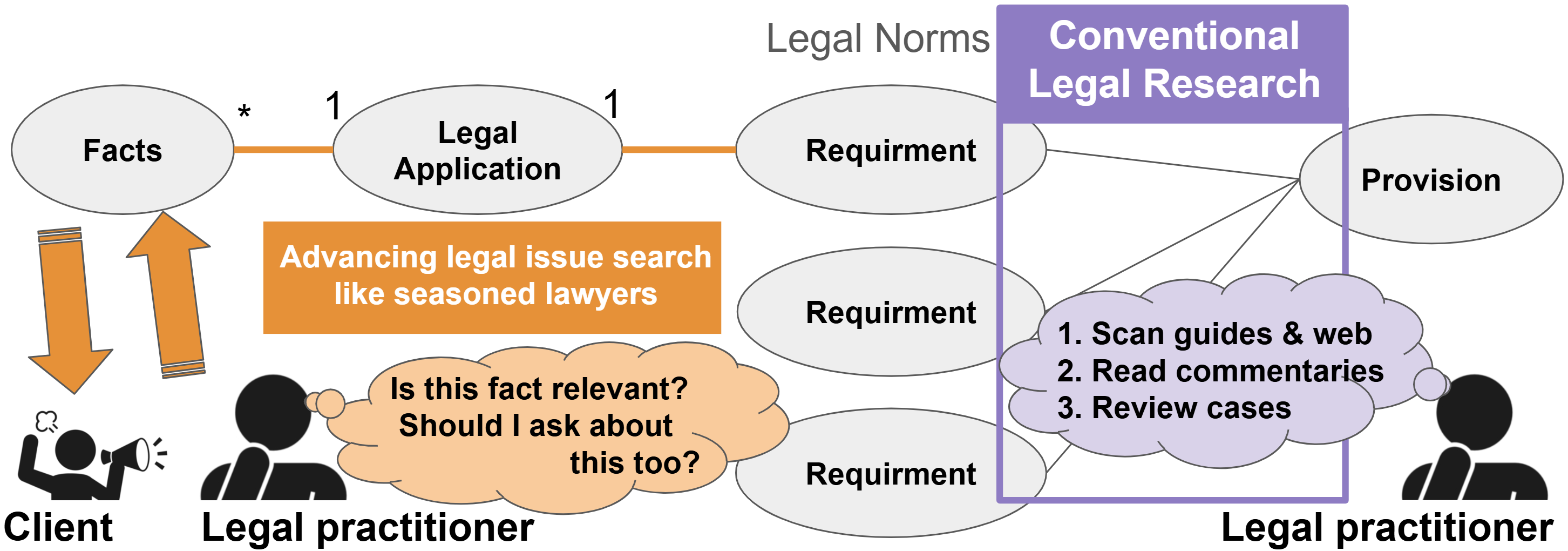}
  \caption{
    Comparison between conventional legal research and our fact-centered retrieval approach.
    Traditional workflows begin with legal provisions and commentaries, relying heavily on expert experience to infer relevant facts.
    Our method reverses this pipeline by starting from observed facts and following structured reasoning paths that connect them to legal norms and provisions.
    This enables issue-spotting behavior, even by junior practitioners, by making the reasoning steps explicit.
  }
  \label{fig:overview}
\end{figure*}

Court judgments provide a uniquely rich source of structured knowledge about how legal reasoning unfolds in practice. Rather than merely stating legal rules, they show how those rules are interpreted and applied to specific factual situations. This reasoning process is shaped by precedent and expert judgment, and reflects patterns that are essential for understanding how legal knowledge is constructed and used. Within each decision lies an implicit, but identifiable, structure that includes evidence, circumstantial facts, primary facts, legal provisions, interpretations, legal applications, and final conclusions~\cite{kondo2024collaborative}.

Specifically, legal reasoning emerges from the interaction of two structured axes: one that builds facts from evidence and one that derives legal meaning from statutes~\cite{kondo2024collaborative,leubsdorf2001structure,sieckmann2020dual,strong2015writing,yao2025elevating}. Along the factual axis, courts identify pieces of evidence, evaluate circumstantial facts, and infer primary facts that support legal evaluation. On the legal axis, legal provisions are interpreted to define the applicable norms and conditions for judgment. These two axes meet at the point of application, where the court determines whether the legal norms apply to the established facts. This intersection defines the core structure of legal inference.

Legal reasoning structures are rarely made explicit. Hence, previous law-focused research in the semantic web community has often overlooked this crucial aspect of legal knowledge. For example, the Legal Knowledge Interchange Format \cite{Hoekstra2007LKIF}, which is a structured framework for representing legal rules and concepts, focuses on static rules, and thus cannot capture the nuanced, context-specific interpretations commonly found in court judgments. Sakhaee and Wilson~\cite{Sakhaee2021LegislationNetwork} focused on dynamic legislation networks, but did not account for the context-dependent reasoning processes that are central to court judgments. Garcia-Godinez et al.~\cite{garcia-godinezDeflationaryApproachLegal2024} emphasized that court judgments are not merely declarations of legal rules, but also illustrate how those rules are interpreted and applied to specific factual situations. However, their work does not provide a formal ontological definition or develop a knowledge graph to represent this structure.

Given the recent rise of large language models (LLMs), it is tempting to ask whether such structured representations are still necessary for the construction of legal search engines. Could a general-purpose LLM be sufficient to answer queries such as ``Given the facts brought before the court, what laws and past precedents should a lawyer or judge consider?''

Although LLMs can generate plausible legal text, they face three fundamental obstacles in supporting reliable legal reasoning. First, they lack the ability to establish legal framing. That is, they cannot determine which jurisdiction, time period, or legal system applies. For example, in defamation cases, whether truth constitutes a defense depends on the jurisdiction. Under US law, truth typically negates unlawfulness, whereas under Japanese law, additional conditions such as public interest and purpose are required~\cite{rodriguez2023japanese}. LLMs often fail to recognize such legal contexts and may produce arguments that conflate incompatible legal standards. Second, LLMs lack the structural grounding needed to connect facts, norms, and provisions through explicit reasoning paths. Although fluent, their outputs are often shallow, mimicking surface-level textual patterns without reconstructing the logic of legal inference. Third, even within a single jurisdiction, courts sometimes adopt multi-layered or divergent perspectives. For instance, in US abortion cases, some lines of judicial opinion support the right, while others oppose it~\cite{Singer2023Conflict}. Simply querying an LLM will not yield a detailed, quantitative summary of how judges construct and weigh these competing legal viewpoints.

Recent studies have addressed these limitations using retrieval-augmented generation (RAG) techniques, which blend LLMs with external document retrieval to anchor the generated answers in authoritative sources~\cite{zheng2024rag-survey}. In the legal domain, ~\cite{chalkidis2024lexfrag} developed a RAG-based system that retrieves supporting fragments from statutes, enabling more accurate and transparent statutory reasoning. However, while these methods improve factual grounding, they do not reconstruct the underlying logic of judicial decision-making or model the structured inferential processes that connect facts, norms, and legal provisions. Systematic evaluation tools such as LegalBench~\cite{NEURIPS2023_89e44582} reveal that, even when assisted with additional context or citation retrieval, LLMs often fail on legal tasks requiring precise, multi-step reasoning and structured context-awareness. These findings suggest that explicit structured representations, such as legal knowledge graphs (LKGs), remain essential for supporting robust and transparent legal search and reasoning systems.

Hence, in this paper, we build on~\cite{kondo2024collaborative} by defining a legal ontology that captures the core reasoning structure found in court decisions. We adopt the resource description framework to encode this graph structure, enabling an explicit and machine-readable representation of legal inference paths. We populate an LKG through a three-step process. First, we extract the key elements of evidence, circumstantial facts, main facts, laws, and legal interpretations from court decisions, refining the prompts used in~\cite{kondo2024collaborative}. Second, we normalize the references to legal provisions so they can be treated as unified nodes. Finally, we create new prompts that link the extracted elements according to the legal ontology, forming structured connections from facts to legal rules. 

Using the constructed LKG, we evaluate performance on a novel class of legal search queries that require factual circumstances to be connected with relevant legal provisions. As shown in Figure~\ref{fig:overview}, conventional legal search engines typically begin with legal provisions and commentaries, and rely heavily on expert intuition to infer the relevant facts. In contrast, our method starts from the observed facts and traces the structured reasoning paths that link them to the applicable legal norms and provisions.

We compare our approach against several baselines, including LLMs and RAG systems that do not incorporate structured knowledge graphs. Our results show that, based on evaluation by expert human judges, the proposed method significantly outperforms these alternatives in both precision and recall.

To summarize, our contributions are as follows:
\begin{itemize}
\item We define a schema for representing structured legal reasoning, capturing how facts, provisions, legal norms, and legal applications are connected within court judgments.
\item We construct an LKG from district-level court decisions in Japanese administrative litigation. The LKG explicitly encodes the factual elements, legal norms, and reasoning paths found in real-world judgments.
\item We develop a novel legal search task and evaluation framework to assess how effectively our system replicates and explains the reasoning patterns observed in expert judicial decisions.
\item All code, data, the legal ontology, and the LKG are available as open-source resources to support future research in legal knowledge modeling and reasoning\footnote{https://github.com/rkondo3/CapturingLegalReasoningPaths}.
\end{itemize}

\section{Method: LKG Construction and Legal Search}
\label{sec:lkg-construction}

Our main methodological contribution is a structured pipeline for constructing an LKG from judgments in Japan. This pipeline captures the structure of legal reasoning. Unlike prior approaches that focus primarily on surface-level legal text or static ontologies, our method explicitly models the inferential relationships among facts, legal norms, legal applications, and provisions. This structured representation enables a fine-grained, interpretable legal search and supports tasks requiring multi-step legal inference. The methodology consists of three main components: (i) schema design, (ii) node and edge extraction for constructing the LKG, and (iii) provision retrieval using the resulting graph.

\subsection{Schema Design}

The LKG schema is designed to capture legal entities and the inferential structure underlying legal reasoning. Many existing ontologies encode legal entities such as statutes and court decisions, but few attempt to model the reasoning process itself, especially the intermediate steps involved in applying norms to facts. Our schema introduces a minimal, but highly expressive, set of classes and properties as a means of representing this structure in a faithful and interpretable way.


The core classes in our RDF-based schema include ``Fact'', ``LegalNorm'', ``LegalApplication'', and ``Provision''. LegalApplication represents an explicit reasoning step that connects a legal norm to a fact. Each LegalApplication node has two outbound RDF properties: ``appliesNorm'', which link to one or more LegalNorms, and ``toFact'', which point to one or more Facts.

\begin{figure}[htbp]
  \centering
  \begin{minipage}{\linewidth}
    \begin{lstlisting}[language=jsonld,label={fig:legalapplication-jsonld}]
...{
  "@id": "LKG:LegalApplication",
  "@type": "owl:Class",
  "rdfs:label": "Legal Application",
  "rdfs:comment": "A reasoning step that applies a legal norm to a fact",
  "rdfs:subClassOf": [
    { "@id": "LKG:LegalNode" },
    { "@id": "schema:Action" }
  ]
},
{
  "@id": "LKG:appliesNorm",
  "@type": "rdf:Property",
  "rdfs:label": "applies norm",
  "rdfs:domain": "LKG:LegalApplication",
  "rdfs:range": "LKG:LegalNorm"
},
{
  "@id": "LKG:toFact",
  "@type": "rdf:Property",
  "rdfs:label": "to fact",
  "rdfs:domain": "LKG:LegalApplication",
  "rdfs:range": "LKG:Fact"
},...
    \end{lstlisting}
  \end{minipage}
  \caption{JSON-LD representation of a legal application and its properties, compatible with the schema.org~\cite{schemaorg} vocabulary.}
  \label{fig:legalapplication-jsonld}
\end{figure}

A key feature of our schema is its ability to represent the hierarchical structure of legal interpretation. Courts often derive higher-level conclusions by combining multiple factual elements. For instance, a person's ``ownership'' of land and their ``physical proximity'' to a loud construction site may together support the conclusion that the person has ``standing'' to file a lawsuit. Similarly, abstract legal norms can be inferred from more specific statutory obligations, such as deriving a ``duty to prevent specific risks'' from a general ``duty to ensure safety''. To support these forms of reasoning, our graph allows connections not only between different categories (such as from LegalNorm to Fact), but also within the same category (such as from Fact to Fact or from LegalNorm to LegalNorm), enabling the modeling of layered legal reasoning.

This explicit modeling of reasoning steps distinguishes our schema from existing legal ontologies, which have traditionally focused on vocabularies and legal references while omitting the logical structure that connects them~\cite{Hoekstra2007LKIF,Sakhaee2021LegislationNetwork}. Our approach treats these inferential steps as first-class elements in the graph, enabling downstream tasks such as a traceable legal search and fine-grained legal reasoning. An illustration of our schema, written in JSON-LD format, is shown in Figure~\ref{fig:legalapplication-jsonld}. The complete schema is available online.

\subsection{Node and Edge Extraction for LKG}

In constructing the LKG, our method for node and edge extraction consists of three steps: (i) preprocessing, (ii) node extraction, and (iii) edge construction. Each step incrementally transforms the raw court judgment text into a structured representation that is aligned with our LKG schema.

\paragraph{\textbf{(i) Preprocessing.}}  

We focus on district court judgments from Japan, using HTML versions provided in the NII legal dataset~\cite{nii-hanrei} (one of the most widely used legal resources in the country). Although the HTML structure appears machine-readable at first glance, it is often inconsistent because of artifacts introduced by automatic generation tools.

The problem of malformed or inconsistent tags and nesting means that we had to manually traverse the HTML node tree and apply regex-based rules to identify and extract meaningful content blocks\footnote{The original PDFs in the dataset are generated via OCR, and the HTML often contains noisy text, such as misaligned punctuation and character misrecognitions; for example, “\begin{CJK}{UTF8}{min}四\end{CJK}” (four) may be misread as “\begin{CJK}{UTF8}{min}匹\end{CJK}” (animal counter). The quality of the NII conversion is not always superior to our own reprocessed PDFs, but we rely on the NII HTML for consistency.}. Headings are detected using a combination of HTML tags, textual features (such as the absence of final punctuation or the presence of initial parentheses), and length-based heuristics. These signals allow the document to be segmented into semantically coherent parts.

In our approach, the section-level structure and source provenance are preserved, as they enable section-aware prompting and maintain traceability to the original document. In particular, we treat the case overview as an independent, structurally significant section, since it is reused in downstream tasks to improve contextual coherence and justification in GPT-based inferences later on.

\begin{figure}[tb]
  \centering
  \begin{tcolorbox}[enhanced, colback=white]
... extract the portions corresponding to the labels below ...

 - Judicial Fact: Case-specific facts derived from submitted evidence or undisputed matters. Do not include legal judgments or interpretations.  
   Example: ``Zalk Noxis was born on Saturn, but currently resides on Mars.''

 - Citation or Reference to Law, Ordinance, Regulation, or Contract: Only the article/paragraph number or cited source itself. Do not include surrounding text.  
   Example: ``Articles 1 and 2 of the Law on Coexistence with Martians.''

 - Legal Norm: The rule presented by law itself. In particular, when the purpose or objective of a law leads to a generalized standard (norm), that portion should be labeled as ``Legal Norm'', even if it explicitly discusses the law’s purpose.  
   Example: ``The Law on Coexistence with Martians aims to promote harmony between humans and extraterrestrial life in light of frequent clashes. Therefore, a `Martian' (Article 2) refers to any non-human lifeform related to Mars.''

 - Legal Application: Application of a law or legal norm to specific facts. When a judgment is made by applying the norm or provision to a concrete situation.  
   Example: ``Because Zalk Noxis currently lives on Mars and is non-human, he qualifies as a `Martian'.''

...

Output must be in JSON format ...

Case Overview:  [CASE OVERVIEW TEXT]

Excerpt from Judgment:   [JUDGMENT TEXT]
  \end{tcolorbox}
\caption{Prompt for node extraction (Japanese original available on GitHub).}
  \label{fig:prompt-node-extraction}
\end{figure}

\paragraph{\textbf{(ii) Node Extraction.}}  

We experimented by extracting graph nodes from segmented court judgments using GPT-4o and Claude Sonnet 3.5. However, Claude did not perform reliably, so we opted to use only GPT-4o. The prompt followed a structured format consisting of role instructions, node category definitions (e.g., \textbf{Fact}, \textbf{Legal Norm}, \textbf{Legal Application}), and prototypical examples. These were carefully aligned with our LKG schema.

To improve consistency, we included a \textit{case overview} at the beginning of the prompt. This helped the model resolve cross-references and interpret the role of each sentence more effectively. We observed a significant improvement in accuracy, particularly in distinguishing between \textbf{Fact} and \textbf{Legal Application} nodes.

Providing few-shot examples typically improves performance~\cite{brown2020language}. However, instead of using real examples, we employed a one-shot fictional scenario based on a fabricated ``Martian Law.'' Because the scenario was fictional, it was easier to detect surface copying, such as the model using the word ``Martian,'' rather than genuine generalization, which is more difficult to assess with real law names. This design also provided two additional benefits: (1) it enabled a controlled demonstration of the desired reasoning process, and (2) it reduced the risk of overfitting to artifacts present in actual legal data. The Martian Law was crafted to reflect edge cases commonly encountered in real court judgments, with guidance from two legal experts.

Finally, statutory reference nodes were re-submitted to the model for normalization. Phrases such as ``the Act'' or ``Article 2'' were standardized into canonical legal titles to enhance downstream linking. For illustrative purposes, a snippet of the prompt is shown in Figure~\ref{fig:prompt-node-extraction}. The full prompt is available online.

\paragraph{\textbf{(iii) Edge Construction.}}

\begin{figure}[tb]
  \centering
  \begin{CJK}{UTF8}{min}
  \begin{tcolorbox}[enhanced, colback=white]
We now provide a list of laws and a list of legal norms.  
For each law, output the corresponding legal norm(s) in JSON format.  
Use laws as keys and legal norms as values.  
Do not output anything other than JSON.

Laws: 
Legal Norms: 
Original excerpt: 

Example:
 
``The Law on Coexistence with Martians...''  
```Martians’ (Article 2) refers...''

Output:  
\{
  "...Article \textbf{1}": "The Law on Coexistence with Martians...",  
  "...Article \textbf{2}": "`Martians' refers to ..."
\}
  \end{tcolorbox}
  \end{CJK}
  \caption{Prompt for linking laws to legal norms.}
  \label{fig:prompt-law-norm}
\end{figure}

\begin{figure}[tb]
  \centering
  \begin{CJK}{UTF8}{min}
  \begin{tcolorbox}[enhanced, colback=white]
Below is a list of legal norms and a sentence of legal application.  
Determine which norm the application is based on and return the most appropriate one.  
Only establish a correspondence if the application clearly and logically relies on a specific norm.  
If no such relationship exists, return an empty list \texttt{[]}.  
Do not link procedural descriptions or mere factual summaries.  
Respond strictly in JSON format; do not use any other formats.

Legal Norms: \\
...\\
\textbf{(28)} Even considering that the noise guidelines ... the train noise in such sections is reasonably assumed to be minimal.

Legal Application:  \\
\textbf{(1)} The plaintiffs qualify as persons who may suffer ... due to the operation of the Nishi--Osaka extension line ...\\
Output:

\{  
  "Application \textbf{1}": ["Norm \textbf{28}"]
\}
  \end{tcolorbox}
  \end{CJK}
  \caption{Prompt for linking legal norms to applications.}
  \label{fig:prompt-norm-application}
\end{figure}

\begin{figure}[tb]
  \centering
  \begin{CJK}{UTF8}{min}
  \begin{tcolorbox}[enhanced, colback=white]
We provide a list of facts and a list of legal application sentences.

A sentence qualifies as a legal application \textbf{only if it satisfies both} of the following conditions:\\
- It is based on explicit legal norms (e.g., statutes, ordinances, ministerial regulations)  
- It legally evaluates concrete facts (e.g., lawful, unlawful)

Do \textbf{not} link the following types of sentences to any facts. Return an empty list \texttt{[]} instead:

- Mere references to technical guidelines or standards  
- Evaluations of party arguments (e.g., “The claim is groundless”)  
- Summaries or organization of facts without legal judgment  
- Explanations of legal frameworks or prerequisites without legal conclusions  
- Descriptions of administrative procedures (e.g., approval by a committee) lacking legal evaluation

Facts:\\
...\\
\textbf{(25)} The plaintiffs reside near the Nishi--Osaka extension line and have filed complaints \texttt{...}

Applications:

\textbf{(1)} The plaintiffs qualify as persons who may suffer \texttt{...} due to the operation of the Nishi--Osaka extension line.

\textbf{(2)} Therefore, Plaintiff 51 is not eligible to bring the lawsuit.

Output:

\{  
  "Application \textbf{1}": ["Fact \textbf{25}"],  
  "Application \textbf{2}": []  
\}
  \end{tcolorbox}
  \end{CJK}
  \caption{Prompt for linking facts to legal applications.}
  \label{fig:prompt-fact-application}
\end{figure}

To capture the legal reasoning paths in court judgments, we again used GPT-based prompts. We focused on three types of directed relations that correspond to stages in legal interpretation:

\begin{itemize}
  \item \textbf{Provision $\rightarrow$ Legal Norm}: capturing interpretive derivation from statutory text
  \item \textbf{Legal Norm $\rightarrow$ Legal Application}: linking abstract norms to legal conclusions
  \item \textbf{Fact $\rightarrow$ Legal Application}: grounding legal conclusions in concrete case facts
\end{itemize}

Language models are subject to input length constraints. Therefore, we partitioned the input contexts and used tailored prompting strategies for each edge type. For \textbf{Provision $\rightarrow$ Legal Norm} edges, we assumed local proximity within sections and paired ``law'' and ``norm'' nodes from the same block. The model, prompted as a legal assistant, returned matching pairs in structured JSON format.

In contrast, ``Legal Norm $\rightarrow$ Legal Application'' and ``Fact $\rightarrow$ Legal Application'' links often span distant sections (sometimes more than eight segments apart). To address this, we adopted a scoped-history prompting strategy: each Legal Application node was paired with all preceding Legal Norm or Fact nodes, allowing the model to reason over the accumulated context. This setup emulates how legal arguments unfold in court judgments, where conclusions frequently depend on reasoning steps distributed across the document.

We validated this design choice by analyzing the section-wise distances between related nodes and found that, in some cases, the distances can be quite large. This finding reinforces the need for a history-aware approach over fixed-size local windows. Although the full analysis is omitted here for brevity, the result supports the modular structure of our edge construction strategy. This modular design enables LLMs to capture multi-step reasoning processes, providing a structured foundation for interpretable and reusable legal inference.

\subsection{Legal Search with LKG}
\label{sec:legal-search}

The final component of our methodology defines legal provision retrieval as the task of identifying relevant statutory articles for a given fact extracted from a court judgment. Our method performs this retrieval by leveraging the LKG described in the previous sections. Each Fact node is embedded using the \texttt{text-embedding-3-large} model and indexed with Annoy~\cite{annoy-github} for an approximate nearest-neighbor search. Embeddings are generated at the level of individual fact sentences, enabling fine-grained retrieval of legally similar cases. For each query fact, the system retrieves the top-$k$ most similar facts and collects the provisions linked to them. To avoid trivial matches, the query fact itself is excluded from retrieval by masking its outgoing edges, referred to as the \textbf{Fact-Masked} setting.

For comparison, we implemented three baselines using GPT 4o. In the \textbf{GPT Simple} setting, the model receives only a single fact sentence and is prompted to return relevant legal provisions. This setup tests whether a pretrained model can identify applicable statutes based on minimal input.

The \textbf{GPT With Context} baseline augments the input with a case overview excerpted from the same judgment. These overviews, which summarize the involved parties and legal issues, are a common feature of Japanese court decisions. By including this context, the model is given a broader legal frame in which to interpret the fact.

The \textbf{GPT With RAG} configuration incorporates RAG. The input fact is embedded using \texttt{text-embedding-3-large}, and semantically similar segments are retrieved from the judgment corpus. Unlike our LKG-based method, which embeds individual fact sentences, this baseline uses \textbf{larger textual units}, such as full judgment sections containing multiple sentences. The retrieved segments are concatenated with the input fact and passed to GPT 4o, allowing the model to generate provision predictions based on retrieved evidence.

To ensure a fair comparison, all baselines were prompted to return relevant statutory provisions in a standardized JSON format. This consistent output format enabled systematic evaluation across all configurations.

\subsection{Dataset}

We focus on judgments issued after 2008, following the introduction of a new document formatting standard in Japans. These judgment documents typically include structured sections such as the main ruling, an overview of the case, and the reasoning of the court. Among these components, we exclusively target the ``reasoning'' section for our analysis because it provides rich textual material connecting facts and legal norms.

We further limit our scope to District Court civil cases, focusing on administrative matters. These judgments offer more detailed factual and legal reasoning than those from higher courts. This choice is driven by practical considerations, because covering all areas of civil law would incur prohibitive computational costs. More importantly, administrative litigation presents favorable conditions for analyzing legal reasoning, typically revolving around the interpretation and application of law rather than fact-intensive disputes.


\section{Evaluation}

This section evaluates whether the proposed LKG effectively supports legal search and retrieval. We focus on three aspects: (1) the accuracy of the extracted graph components, (2) whether the resulting knowledge graph exhibits a coherent structure, and (3) how well the LKG enables provision retrieval from factual inputs.

\subsection{Extraction Accuracy}

\begin{table}[tb]
  \centering
  \caption{Evaluation of GPT with designed prompts for node extraction in LKG construction, as assessed by legal experts}
  \label{tab:node-extraction-eval}
  \begin{tabular}{p{2.2cm} r r r r r r}
    \hline
    Node Type & TP & FP & FN & \makecell[l]{Precision} & \makecell[l]{Recall} & F1 \\
    \hline
    Provision           & 119 & 0  & 0  & 1.0000 & 1.0000 & 1.0000 \\
    Norm      & 64  & 8  & 1  & 0.8889 & 0.9846 & 0.9347 \\
    Application     & 60  & 0  & 9  & 1.0000 & 0.8696 & 0.9302 \\
    Fact            & 193 & 1  & 1  & 0.9948 & 0.9948 & 0.9948 \\
    \hline
  \end{tabular}
\end{table}

We evaluated the accuracy of GPT's node and edge extractions, prompted using our designed instructions, through a manual annotation study conducted with two legal experts. The experts reviewed 57 randomly selected sections of court judgments and annotated nodes corresponding to the four categories in our LKG: \textbf{Fact}, \textbf{Provision}, \textbf{Legal Norm}, and \textbf{Legal Application}. These categories follow the same definitions used in our prompt design: Fact refers to case-specific factual findings without legal interpretation; Provision refers to explicit references to statutes, regulations, or contracts; Legal Norm refers to normative propositions derived from the law; and Legal Application refers to the application of a legal norm or law to a specific fact.

Each GPT output was compared with the human annotations, and we recorded the numbers of true positives (TPs), false positives (FPs), and false negatives (FNs) per category. Table~\ref{tab:node-extraction-eval} indicates that, overall, the model achieved high precision and recall, with perfect scores for Provision and near-perfect performance for Fact. Recall was slightly lower for Legal Application, primarily because of frequent misclassification as Legal Norm. In these cases, segments that should have been identified as Legal Application were instead labeled as Legal Norm by the model. This error pattern aligns with our prompt design, which reflects the domain knowledge that Legal Application generally presupposes the presence of a corresponding Legal Norm. Therefore, we do not regard this behavior as an extraction failure, but consider it consistent with the structure the prompt was designed to elicit.

After evaluating node extraction, we proceeded to assess the quality of edge extraction. The legal experts manually reviewed the predicted edges across 92 randomly selected judgment sections. For each edge, they assigned one of three labels: TP, FP, or FN. An edge was labeled as TP if it correctly connected the intended pair of nodes with the appropriate relation type. Edges that linked incorrect node pairs or used the wrong relation type were labeled as FP, while necessary edges that had been omitted were labeled as FN.

\begin{table}[tb]
  \centering
  \caption{Evaluation of GPT with designed prompts for edge extraction in LKG construction, as assessed by legal experts}
  \label{tab:edge-extraction-eval}
  \begin{tabular}{p{2.2cm} r r r r r r}
    \hline
    Edge Type & TP & FP & FN & \makecell[l]{Precision} & \makecell[l]{Recall} & F1 \\
    \hline
    Provision $\rightarrow$ Norm     & 135 & 8  & 7  & 0.9434 & 0.9507 & 0.9470 \\
    Norm $\rightarrow$ Application & 78  & 18 & 5  & 0.8125 & 0.9398 & 0.8710 \\
    Application $\rightarrow$ Fact     & 105 & 5  & 8  & 0.9545 & 0.9292 & 0.9417 \\
    \hline
  \end{tabular}
\end{table}

According to Table~\ref{tab:edge-extraction-eval}, GPT demonstrated strong overall performance in edge extraction, achieving precision above 0.94 for most edge types. In the case of \textit{Legal Norm} $\rightarrow$ \textit{Legal Application}, a higher number of FP cases occurred. These were mainly the result of two factors: overextension of \textit{Legal Norm} spans, and missing links to supporting norms in layered legal reasoning structures. For example, GPT often captured the primary norm applied to a fact, but failed to link to additional norms that support or constrain the main one. This limitation reflects a structural constraint in our current schema, which does not support edges between norms (e.g., \textit{Legal Norm} $\rightarrow$ \textit{Legal Norm}).

From the perspective of a legal retrieval system, minimizing omissions and ensuring dense graph connectivity are important. In this respect, the model’s output is largely desirable. In particular, the capture of hierarchical reasoning structures in which legal norms and facts are layered across multiple levels will be enhanced by expanding the edge schema to support more nuanced connections.

\subsection{Structural Analysis of the LKG}

\begin{table}[tb]
  \centering
  \caption{Graph statistics by node class}
  \label{tab:node-stats}
  \begin{tabular}{p{1.5cm} r r r r}
    \hline
    \makecell[l]{Node Type} & \# Nodes & \makecell[l]{Average\\In-Degree} & \makecell[l]{Average\\Out-Degree} & \makecell[l]{\# Self-Loops} \\
    \hline
    Provision         & 10,552 & 0.00 & 1.70 & 0 \\
    Norm        & 15,445 & 1.17 & 0.95 & 25 \\
    Application       & 8,957  & 3.80 & 0.13 & 746 \\
    Fact              & 11,199 & 0.18 & 1.70 & 472 \\
    \hline
  \end{tabular}
\end{table}

\begin{table}[tb]
  \centering
  \caption{Graph statistics by edge type}
  \label{tab:edge-stats}
  \begin{tabular}{p{5.1cm} r r}
    \hline
    \makecell[l]{Edge Type\\(Source $\rightarrow$ Target)} & \makecell[c]{\# Edges} & \makecell[l]{Average\\Multiplicity} \\
    \hline
    Provision $\rightarrow$ Norm          & 18,814 & 1.69 \\
    Norm $\rightarrow$ Application  & 15,356 & 2.20 \\
    Fact $\rightarrow$ Application        & 19,242 & 1.63 \\
    \hline
  \end{tabular}
\end{table}

Subsequently, we assess whether the overall structure of the constructed LKG reflects desirable properties for downstream legal reasoning. Specifically, we analyze whether the graph exhibits hierarchical composition, element reuse, and localized modularity. Tables~\ref{tab:node-stats}, \ref{tab:edge-stats}, and~\ref{tab:network-stats} provide the relevant statistics.

At the node level, \textit{Legal Application} nodes tend to have high in-degree and low out-degree, consistent with their role as endpoints of reasoning. \textit{Provision} and \textit{Fact} nodes more often serve as sources. \textit{Legal Norm} nodes act as intermediaries. Self-loops occur primarily in \textit{Fact} and \textit{Application} nodes, suggesting that certain segments fulfill multiple roles.

At the edge level, we observe frequent reuse of \textit{Fact} and \textit{Norm} nodes across applications. This pattern indicates that conclusions are often supported by multiple inputs and that key legal elements participate in multiple reasoning chains.

At the global level, the graph is sparse and fragmented, consisting of many weakly connected components. This structure reflects the document-level nature of construction: each judgment forms an isolated subgraph. The graph’s lack of strong connectivity is consistent with the forward-only, acyclic flow of legal logic.


\begin{table}[tb]
  \centering
  \caption{Graph statistics at the network level. WCC stands for weakly connected component.}
  \label{tab:network-stats}
  \begin{tabular}{p{6.5cm} r}
    \hline
    Metric & Value \\
    \hline
    Number of Nodes & 44,447 \\
    Number of Edges & 51,296 \\
    Number of WCCs with diameter $\geq 2$ & 1,478 \\
    Avg. Diameter of WCCs ($\geq 2$) & 3.10 \\
    Std. Dev. of Diameter & 2.39 \\
    Std. Dev. of Degree & 5.28 \\
    Graph Density & $2.59 \times 10^{-5}$ \\
    \hline
  \end{tabular}
\end{table}


\subsection{Evaluation as Legal Search Engine}
\label{sec:evaluation-search}

\begin{table*}[t]
  \centering
  \caption{Evaluation results for GPT baselines and LKG retrieval ($k=1$-$7$). 
  The total number of gold-standard labels is 1242.
Bold indicates the best value in each column. Underline indicates GPT outputs and $k$ settings in LKG retrieval that yield a similar number of outputs.}
    \label{tab:legal-search}
  \begin{tabular}{p{4cm}rrrrrrrrr}
    \toprule
    Method & Pred & TP & \shortstack{Macro\\Recall} & \shortstack{Micro\\Recall} & \shortstack{Macro\\Precision} & \shortstack{Micro\\Precision} & \shortstack{Macro\\F1} & \shortstack{Micro\\F1} \\
    \midrule
    GPT Simple & 4922 & 123 & 0.046 & 0.099 & 0.037 & 0.025 & 0.041 & 0.040 \\
    GPT With Context (baseline) & \underline{7377} & 336 & 0.137 & 0.271 & 0.054 & 0.046 & 0.077 & 0.078 \\
    
    GPT With RAG & \underline{7792} & 436 & 0.176 & 0.351 & 0.062 & 0.056 & 0.092 & 0.097 \\
    \hdashline
    LKG Retrieval ($k=1$) & 1253 & 497 & 0.203 & 0.400 & 0.199 & \textbf{0.397} & 0.201 & \textbf{0.398} \\
    LKG Retrieval ($k=2$) & 2403 & 712 & 0.278 & 0.573 & \textbf{0.235} & 0.296 & 0.255 & 0.391 \\
    LKG Retrieval ($k=3$) & 3470 & 829 & 0.311 & 0.667 & 0.227 & 0.239 & \textbf{0.263} & 0.352 \\
    LKG Retrieval ($k=4$) & 4415 & 887 & 0.330 & 0.714 & 0.213 & 0.201 & 0.259 & 0.314 \\
    LKG Retrieval ($k=5$) & 5423 & 920 & 0.340 & 0.741 & 0.194 & 0.170 & 0.247 & 0.276 \\
    LKG Retrieval ($k=6$) & 6464 & 953 & 0.354 & 0.767 & 0.179 & 0.147 & 0.238 & 0.247 \\
    LKG Retrieval ($k=7$) & \underline{7482} & \textbf{965 }& \textbf{0.359 }& \textbf{0.777} & 0.161 & 0.129 & 0.222 & 0.221 \\
    \bottomrule
  \end{tabular}
\end{table*}

Finally, we evaluate the effectiveness of LKG-based retrieval using both quantitative and qualitative methods. First, we compare retrieval accuracy against GPT-based baselines. Then, we analyze representative reasoning paths to assess whether the system captures judicial logic beyond surface-level matching.
We evaluated LKG-based retrieval in comparison with three GPT-4o-based baselines introduced in Section~\ref{sec:legal-search}. The benchmark dataset was derived from Japanese administrative litigation judgments, where each fact sentence was linked to gold-standard provisions via the LKG.


Table~\ref{tab:legal-search} compares performance across methods. We report predicted provisions (Pred), true positives (TPs), and the resulting precision, recall, and F1 scores under both macro and micro evaluation metrics. Macro scores treat each case equally by averaging over judgments, whereas micro scores aggregate globally across predictions.
\textbf{LKG Retrieval under the Fact-Masked setting significantly outperforms all GPT-based methods in both macro and micro recall.} For example, at $k=3$, the macro recall is 0.311 and the micro recall is 0.667, far exceeding GPT With Context (0.137 and 0.271, respectively). Although $k=2$ yields slightly higher micro recall (0.671), $k=3$ provides better macro recall and more stable performance. Full results for $k = 1$--$7$ are provided in the appendix.
Among GPT baselines, GPT With Context outperforms GPT Simple, highlighting the value of case overviews. However, \textbf{GPT With RAG fails to bridge the gap with LKG Retrieval}, despite access to retrieved context. This suggests that unstructured retrieval lacks the structural alignment required for multi-step legal reasoning.

While these quantitative results validate the overall performance of LKG-based retrieval, they do not reveal how well the system captures the logic of legal reasoning. To address this, we turn to qualitative evaluation. We examine complete reasoning paths extracted by the system and assess whether they faithfully mirror judicial interpretive structure. The following examples demonstrate how the LKG supports both straightforward and nuanced inference patterns.


We begin with a simple case that the system correctly extracted without manual correction. The court addressed a resident audit request governed by \emph{Article~242 of the Local Autonomy Act}. It examined a \textbf{Fact} where \emph{the request was filed on June 1, 2007}, and concluded in its \textbf{Application} that \emph{the request was inadmissible due to expiration of the statutory deadline}. This conclusion rested on a \textbf{Norm} specifying that \emph{the deadline must be calculated from the date of the boundary confirmation consultation or its completion}. This example shows that the LKG correctly linked the fact, norm, and statutory provision to form a legally valid path.

More complex examples reveal how factual variation can yield different legal outcomes even under the same norm and provision. In urban planning disputes, one case presented a \textbf{Fact} where \emph{the plaintiffs lived, worked, or owned property near the project site and claimed harm to their environment and property use}. The court held that they had standing to seek cancellation, applying a \textbf{Norm} that \emph{only those with a legally protected interest may file such a suit}, grounded in \emph{Article~9 of the Administrative Case Litigation Act}.
We also identified a contrasting case in which the plaintiff resided \emph{556 meters from the site}, a distance the court found insufficiently direct, leading to an \textbf{Application} denying standing. These cases illustrate how minor factual differences can materially affect legal reasoning.

Finally, we highlight a different type of variation, where legal norms interact with constitutional principles. In one case, the \textbf{Fact} was that \emph{the plaintiff acquired the nationality of Country~A through a document submitted by their parents, which met the legal requirements under the former nationality law of Country~A}. The court deemed this acquisition voluntary and applied the \textbf{Norm} that \emph{a person who voluntarily acquires a foreign nationality loses Japanese nationality}, pursuant to \emph{Article~11 of the Nationality Act}. 
The LKG-based retrieval system also surfaced a structurally corresponding case in which the same norm and provision were challenged on constitutional grounds. In that case, the \textbf{Fact} was that \emph{the plaintiffs expressed their intent to acquire the nationalities of Country~B and Country~C, and filed lawsuits claiming that Article~11 violated their freedom to change nationality}. The court rejected the claim, applying the \textbf{Application} that \emph{the provision did not exceed legislative discretion and was consistent with the constitutional guarantee of nationality renunciation}, affirming the statute’s aim of preventing dual nationality.

\section{Conclusion}
We have introduced a structured LKG that captures the multi-step reasoning paths embedded in Japanese administrative court judgments. By modeling the connections between facts, legal norms, and legal applications, our LKG makes the legal reasoning explicit and machine-interpretable. To evaluate the utility of this LKG, we designed a legal provision retrieval task and compared the performance of GPT- and graph-based methods. The results show that structured legal knowledge enables more accurate and consistent reasoning than language models alone, even when they have access to relevant source content. These findings demonstrate that legal AI systems benefit from explicitly grounded knowledge structures, reaffirming the essential role of LKGs in interpretable and generalizable legal reasoning.

\section{Acknowledgements}
We thank TKC Corporation for helpful discussions. This work uses judicial precedent data from the NII Informatics Research Data Repository. R.K. is funded by JST, ACT-X JPMJAX23CA. R.H. is supported by the JST FOREST Program (JPMJFR216Q), JST PRESTO Program (JPMJPR2469), a Grant-in-Aid for Scientific Research (KAKENHI) (JP24K03043), and the UTEC-UTokyo FSI Research Grant Program. 


\bibliographystyle{plainnat}
\bibliography{LegalSearch2025}

\end{document}